# FedKBP: Federated dose prediction framework for knowledge-based planning in radiation therapy

Jingyun Chen*[a], Martin King[b], Yading Yuan[a,c]
[a]Department of Radiation Oncology, Columbia University Irving Medical Center
622 W 168th St, New York, NY 10032, United States
[b]Department of Radiation Oncology, Brigham and Women's Hospital, Harvard Medical School
75 Francis St, Boston, MA 02445, United States
[c]Data Science Institute, Columbia University
*Email: jc6171@cumc.columbia.edu

## ABSTRACT

Dose prediction plays a key role in knowledge-based planning (KBP) by automatically generating patient-specific dose distribution. Recent advances in deep learning-based dose prediction methods necessitates collaboration among data contributors for improved performance. Federated learning (FL) has emerged as a solution, enabling medical centers to jointly train deep-learning models without compromising patient data privacy. We developed the FedKBP framework to evaluate the performances of centralized, federated, and individual (i.e. separated) training of dose prediction model on the 340 plans from OpenKBP dataset. To simulate FL and individual training, we divided the data into 8 training sites. To evaluate the effect of inter-site data variation on model training, we implemented two types of case distributions: 1) Independent and identically distributed (IID), where the training and validating cases were evenly divided among the 8 sites, and 2) non-IID, where some sites have more cases than others. The results show FL consistently outperforms individual training on both model optimization speed and out-of-sample testing scores, highlighting the advantage of FL over individual training. Under IID data division, FL shows comparable performance to centralized training, underscoring FL as a promising alternative to traditional pooled-data training. Under non-IID division, larger sites outperformed smaller sites by up to 19% on testing scores, confirming the need of collaboration among data owners to achieve better prediction accuracy. Meanwhile, non-IID FL showed reduced performance as compared to IID FL, posing the need for more sophisticated FL method beyond mere model averaging to handle data variation among participating sites.

## 1. INTRODUCTION

To address the growing demand for radiation therapy in cancer treatment, knowledge-based planning (KBP) has been introduced to streamline the planning process and reduce treatment lead time[1]. Central to KBP is the dose prediction, which automatically estimates patient-specific dose distribution for treatment plan evaluation and optimization. Recent advances in deep learning-based dose prediction methods[2,3,4,5] have highlighted the challenge of limited training data availability[5], necessitating collaboration among diverse data contributors. Federated learning (FL)[6] has emerged as a solution, enabling medical centers to jointly train deep-learning models without compromising patient data privacy (Figure 1). To evaluate the effectiveness of federated dose prediction in KBP, we developed a framework named FedKBP to access the performances of centralized, federated, and individual (i.e. separated) training of dose prediction model on same dataset.

## 2. MATERIALS AND METHODS

### Data

This study employed the public dataset from open-access knowledge-based planning (OpenKBP) grand challenge[7]. The OpenKBP dataset contains a total of 340 cases (i.e. plans) for head and neck cancer treatment, including 200 training cases, 40 validating cases, and 100 testing cases. The prescribed plans were 70, 63 and 56 Gy in 35 fractions. The organs at risk (OARs) include brain stem, right parotid, left parotid, spinal cord, larynx, mandible and esophagus. To simulate the distributed training environment, we randomly divided the 200 training cases and 40 validating cases into 8 groups, representing 8 training sites. To evaluate the effect of inter-site data variation on model training, we implemented two types of case distributions: 1) Independent and identically distributed (IID), where the training and validating cases were evenly divided among the 8 sites, and 2) non-IID, where some sites have more cases than others. Table 1 showed the case numbers of each site under IID and non-IID data distributions. The 100 testing cases were used as out-of-sample testing cases by each site, regardless of data distribution type.

Table 1. Number of training and validating cases of each site in FedAvg and IM

| Distribution | Site | Training cases | Validating cases |
|---|---|---|---|
| IID | Site0 | 25 | 5 |
| | Site1 | 25 | 5 |
| | Site2 | 25 | 5 |
| | Site3 | 25 | 5 |
| | Site4 | 25 | 5 |
| | Site5 | 25 | 5 |
| | Site6 | 25 | 5 |
| | Site7 | 25 | 5 |
| non-IID | Site0 | 40 | 8 |
| | Site1 | 35 | 7 |
| | Site2 | 30 | 6 |
| | Site3 | 25 | 5 |
| | Site4 | 25 | 5 |
| | Site5 | 20 | 4 |
| | Site6 | 15 | 3 |
| | Site7 | 10 | 2 |

**Dose prediction model**

In this study, we implemented the Scale Attention Network (SANet) for 3D dose prediction, with voxel-wise Mean Absolute Error (MAE) between dose prediction and ground truth as the loss function[2,8]. SANet features a dynamic scale attention mechanism to incorporate low-lever details with high-level semantics from feature maps at different scales, thus, to better integrate information across different scales. It has demonstrated strong performance in dose prediction[8] as well as tumor segmentation[9,10].

**Baselines and evaluation**

We evaluated three different training methods: 1) Pooled model (PM), which is trained with all 8 sites' training data together. 2) Individual model (IM), which is trained on each site trains with its local data individually without exchanging models. 3) FedAvg: which forms a consensus model at each round by weighted average of the local models from all sites[6] (Figure 1). For FedAvg and IM, we further evaluated the two types of data distributions (IID and non-IID). The resulting model performances are accessed with dose score and dose-volume histogram (DVH) score as defined in OpenKBP challenge[7]. For both scores, smaller value represents higher prediction accuracy and better model performance.

**Implementation**

All three training methods (PM, FedAvg, IM) were carried out with in-house developed FedKBP framework on Nvidia GTX 1080 TI GPUs with 11 GB memory. Each method was trained for 100 epochs. For FedAvg and IM, each site was assigned a separate GPU for training.

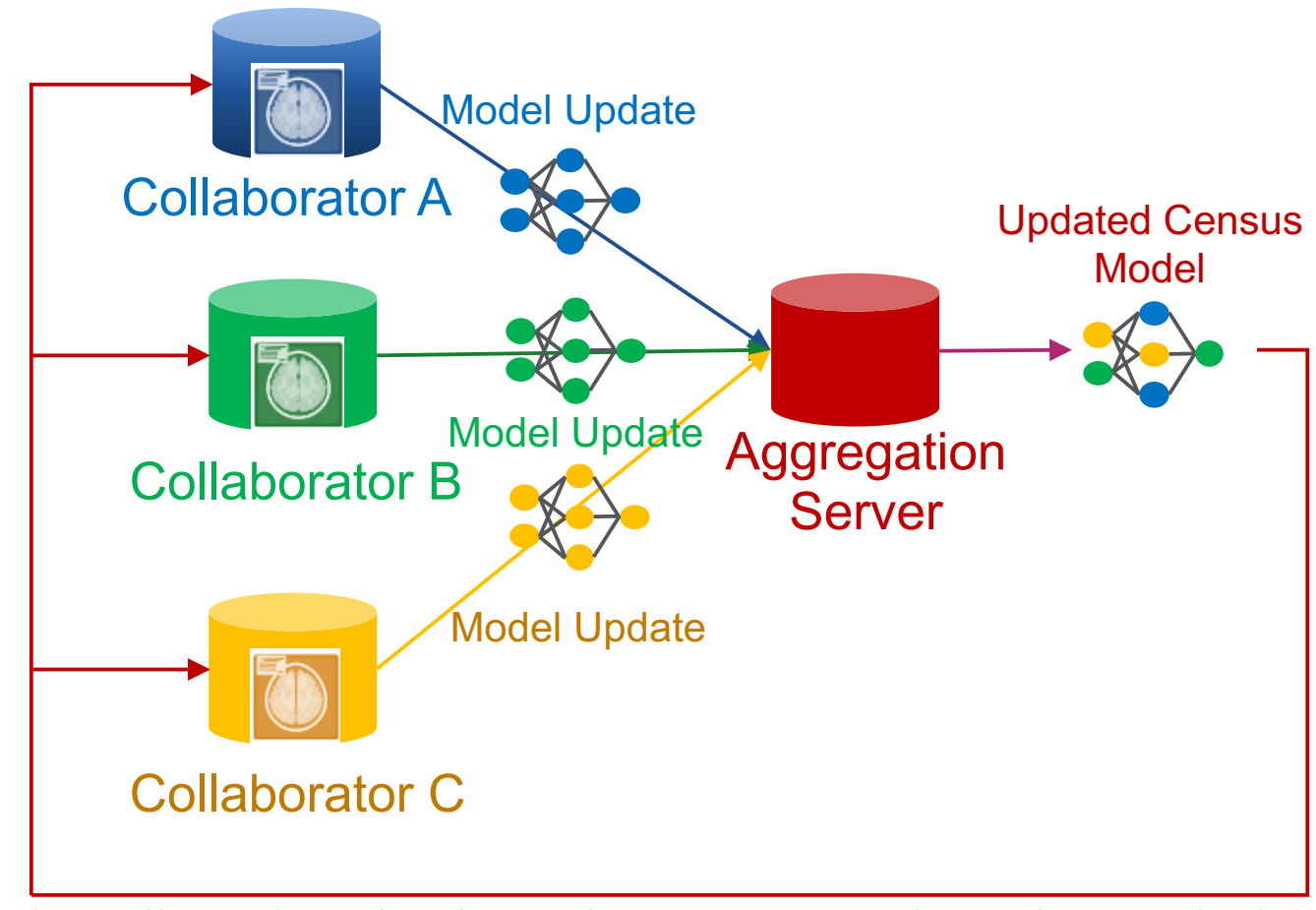

Fig 1. Illustration of FedAvg, the most commonly used FL method

## 3. RESULTS

Table 2 contains the model prediction accuracy (in terms of mean dose and DVH scores) for different training scenarios, based on the testing cases. For PM and FedAvg, the corresponding global models are evaluated. For IM, the testing scores of local models are computed and averaged to represent the global performance.

Table 2. Mean dose and DVH scores for different training scenarios

| | **Methods** | **Mean Dose score (↓)** | **Mean DVH score (↓)** |
|---|---|---|---|
| Global/Average | PM | 2.6758 | 1.8990 |
| | FedAvg IID | 2.7254 | 1.8530 |
| | FedAvg non-IID | 2.9001 | 1.9979 |
| | IM IID | 3.4288 | 2.6816 |
| | IM non-IID | 3.4528 | 2.5843 |
| Local/Individual | Site0 IID | 3.4378 | 2.3753 |
| | Site1 IID | 3.3847 | 2.5776 |
| | Site2 IID | 3.5069 | 3.0884 |
| | Site3 IID | 3.3451 | 2.3336 |
| | Site4 IID | 3.4927 | 3.0827 |
| | Site5 IID | 3.3765 | 2.6797 |
| | Site6 IID | 3.4387 | 2.5786 |
| | Site7 IID | 3.4480 | 2.7369 |
| | Site0 non-IID | 3.1687 | 2.3872 |
| | Site1 non-IID | 3.2214 | 2.1145 |
| | Site2 non-IID | 3.2722 | 2.3520 |
| | Site3 non-IID | 3.5199 | 2.6569 |
| | Site4 non-IID | 3.4691 | 2.6622 |
| | Site5 non-IID | 3.5274 | 2.8111 |
| | Site6 non-IID | 3.6877 | 2.8457 |
| | Site7 non-IID | 3.7557 | 2.8450 |

On global/average level, PM showed high performance on both dose and DVH scores as expected, since it was trained on all the data pooled together. FedAvg consistently outperformed IM on both testing scores (dose and DVH) and both data divisions (IID and non-IID), showing prevalent advantages of FedAvg over IM. Notably, the IID FedAvg showed comparable performance to PM, with larger dose score (2.7254 vs 2.6758) but smaller DVH score (1.8530 vs 1.8990). This indicates FL can potentially achieve similar performance as compared to pooled training, even though the former is trained on distributed data. As compared to IID FedAvg, non-IID FedAvg suffered with larger testing scores, for both dose (2.9001 vs 2.7254) and DVH (1.9979 vs 1.8530). This finding confirms the challenge of non-IID data in FL, which is consistent with reports from previous publications[11,12].

On local/individual level, the IID IM showed coherent performances among local sites, while exhibiting some individual variability. For the non-IID IM, larger sites (i.e. sites with more cases) tend to have better performance than smaller sites. For example, the largest non-IID site (Site0, with 40 training and 8 validating cases) outperformed the smallest non-IID site (Site7, with 10 training cases and 2 validating cases) by 18% on dose score (3.1687 vs 3.7557) and 19% on DVH score (2.3872 vs 2.8450). This finding shows the benefit of large dataset and confirms the necessity of collaboration among data contributors for improved performance.

Figure 2 shows the validation loss over epoch (based on the validating cases) representing the model optimization over time under different training scenarios. For FedAvg and IM, the validation losses of local sites are averaged to compare with the global validation loss of PM. As expected, the validation loss of PM displays fastest decreases and lowest final value among all training scenarios. Empowered by model exchange at each epoch, the validation loss of FedAvg showed faster decrease and lower final values than IM under both IID and non-IID data distributions, reflecting the enhanced model optimization in FedAvg over IM. In both FedAvg and IM, the non-IID group are consistently outperformed by the IID counterpart, confirming the hindrance to model optimization by non-IID data distributions.

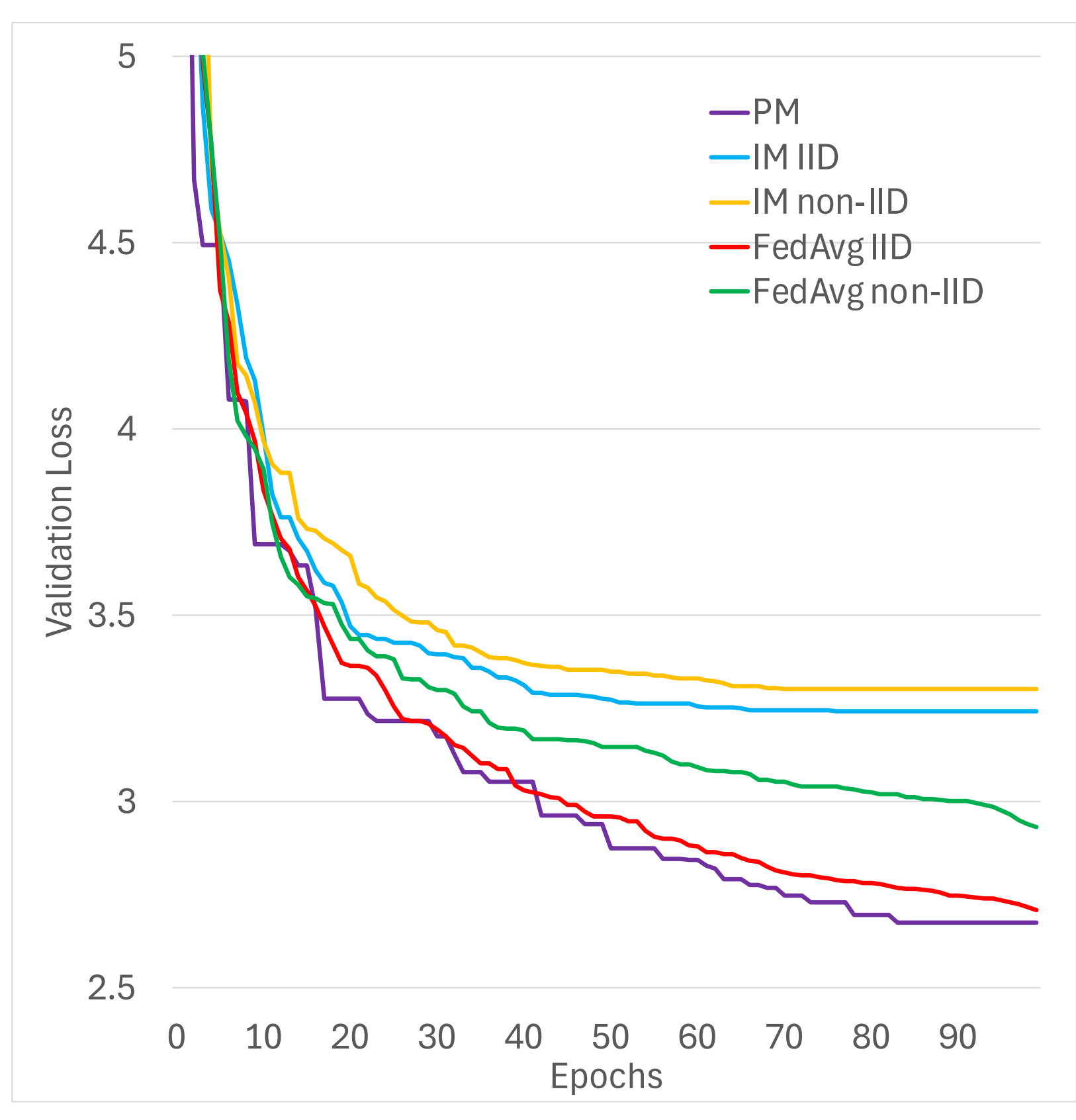

Fig 2. Validation loss over epochs for different training scenarios

## 4. DISCUSSION

We developed a novel framework, namely FedKBP, for simulating and benchmarking various training scenarios of 3D dose prediction models. Under FedKBP framework, we accessed the dose prediction models from three distinctive training approaches under IID and non-IID data distributions. The results underscored FL as a promising solution for training dose prediction model over distributed data, while also confirmed that non-IID data distribution poses a challenge to FL.

One limitation of this study stems from the OpenKBP dataset. While the dataset was collected from multiple contributing centers, the data underwent cleaning and harmonization at a single center, resulting in a consistent dataset that may not fully reflect the heterogeneous distributions observed in real-world scenarios. Furthermore, due to the absence of source center information in the OpenKBP dataset [7], we implemented a random data division with arbitrary case assignments. To emulate a non-IID data distribution, we allocated data to sites with varying sample sizes. Consequently, the limited overall size of the OpenKBP dataset resulted in small sample sizes for individual sites like non-IID Site 6 and 7 in Table 1, potentially leading to insufficient training of deep learning models. For future work, we aim to utilize a larger dataset with source center information to enable more realistic site-specific data division.

## 5. CONCLUSIONS

Federated learning (FL) offers a promising alternative to centralized data-pooling, delivering comparable performance while preserving data privacy at local sites. However, non-IID data distribution across sites may hinder model optimization in FL. Therefore, more sophisticated and personalized FL method is needed to tackle data variations among participating sites and achieve enhanced performance.

## ACKNOWLEDGEMENT

This work is supported by a research grant from Varian Medical Systems (Palo Alto, CA, USA), UL1TR001433 from the National Center for Advancing Translational Sciences, and R21EB030209 from the National Institute of Biomedical Imaging and Bioengineering of the National Institutes of Health, USA. The content is solely the responsibility of the authors and does not necessarily represent the official views of the National Institutes of Health. This research has been partially funded through the generous support of Herbert and Florence Irving/the Irving Trust.

## REFERENCES

[1] Wu, B., Ricchetti, F., Sanguineti, G., Kazhdan, M., Simari, P., Chuang, M., Taylor, R., Jacques, R. and McNutt, T., "Patient geometry-driven information retrieval for IMRT treatment plan quality control," Med Phys 36(12), 5497–5505 (2009).

[2] Yuan, Y., Tseng, T., Chao, M., Stock, R. and Lo, Y., "Predicting three dimensional dose distribution with deep convolutional neural networks," Med. Phys 44(6) (2017).

[3] Yuan, Y., Tseng, T.-C. and Lo, Y.-C., "3d deep planning radiotherapy system and method," WO2019027924A1 (2018).

[4] Nguyen, D., Long, T., Jia, X., Lu, W., Gu, X., Iqbal, Z. and Jiang, S., "A feasibility study for predicting optimal radiation therapy dose distributions of prostate cancer patients from patient anatomy using deep learning," Sci Rep 9(1), 1076 (2019).

[5] Liu, S., Zhang, J., Li, T., Yan, H. and Liu, J., "Technical Note: A cascade 3D U-Net for dose prediction in radiotherapy," Med Phys 48(9), 5574–5582 (2021).

[6] McMahan, B., Moore, E., Ramage, D., Hampson, S. and Arcas, B. A. y., "Communication-Efficient Learning of Deep Networks from Decentralized Data," Proceedings of the 20th International Conference on Artificial Intelligence and Statistics 54, A. Singh and J. Zhu, Eds., 1273–1282, PMLR (2017).

[7] Babier, A., Zhang, B., Mahmood, R., Moore, K. L., Purdie, T. G., McNiven, A. L. and Chan, T. C. Y., "OpenKBP: The open-access knowledge-based planning grand challenge and dataset," Med Phys 48(9), 5549–5561 (2021).

[8] Adabi, S., Tzeng, T.-C. and Yuan, Y., "Predicting 3D dose distribution with scale attention network for prostate cancer radiotherapy," Medical Imaging 2022: Image-Guided Procedures, Robotic Interventions, and Modeling, C. A. Linte and J. H. Siewerdsen, Eds., 40, SPIE (2022).

[9] Yuan, Y., "Automatic Head and Neck Tumor Segmentation in PET/CT with Scale Attention Network," Head and Neck Tumor Segmentation, V. Andrearczyk, V. Oreiller, and A. Depeursinge, Eds., 44–52, Springer International Publishing, Cham (2021).

[10] Yuan, Y., "Evaluating Scale Attention Network for Automatic Brain Tumor Segmentation with Large Multi-parametric MRI Database," [International MICCAI Brainlesion workshop], 42–53 (2022).

[11] Shoham, N., Avidor, T., Keren, A., Israel, N., Benditkis, D., Mor-Yosef, L. and Zeitak, I., "Overcoming Forgetting in Federated Learning on Non-IID Data" (2019).

[12] Gao, Z., Wu, F., Gao, W. and Zhuang, X., "A New Framework of Swarm Learning Consolidating Knowledge from Multi-Center Non-IID Data for Medical Image Segmentation," IEEE Trans Med Imaging 42(7), 2118–2129 (2023).